\documentclass{ecai}

\usepackage{graphicx}
\usepackage{booktabs}
\usepackage{latexsym}
\usepackage{multirow}
\usepackage{balance}
\usepackage{xurl}

\usepackage{graphicx}
\usepackage{multicol}
\usepackage{relsize}
\usepackage{listings} 

\usepackage{amsmath}
\usepackage{amssymb}
\usepackage{amsfonts}
\usepackage{algorithm}
\usepackage{algpseudocode}
\usepackage[hidelinks]{hyperref}
\usepackage{orcidlink}

\newcommand{\R}{\mathbb{R}}

\newcommand{\T}{\mathcal{T}}
\newcommand{\A}{\mathcal{A}}

\newtheorem{definition}{Definition}
\DeclareMathOperator*{\argmax}{arg\,max}

\begin{document}

\begin{frontmatter}

\title{Algorithms for learning value-aligned policies considering admissibility relaxation}

\author[A]{\fnms{Andrés}~\snm{Holgado-Sánchez}\orcid{0000-0001-8853-1022}\orcidlink{0000-0001-8853-1022}}
\author[A]{\fnms{Joaquín}~\snm{Arias}\orcid{0000-0003-4148-311X}\orcidlink{0000-0003-4148-311X}}
\author[A]{\fnms{Holger}~\snm{Billhardt}\orcid{0000-0001-8298-4178}\orcidlink{0000-0001-8298-4178}}
\author[A]{\fnms{Sascha}~\snm{Ossowski}\orcidlink{0000-0003-2483-9508}\orcid{0000-0003-2483-9508}}

\address[A]{CETINIA, Universidad Rey Juan Carlos, Madrid, Spain\\
  \{andres.holgado, joaquin.arias, holger.billhardt, sascha.ossowski\}@urjc.es}
  
\begin{abstract}
\hyphenpenalty 10000
The emerging field of \emph{value awareness engineering} claims that software agents and systems should be value-aware, i.e. they must make decisions in accordance with human values. In this context, such agents must be capable of explicitly reasoning as to how far different courses of action are aligned with these values. For this purpose, values are often modelled as preferences over states or actions, which are then aggregated to determine the sequences of actions that are maximally aligned with a certain value. Recently, additional value admissibility constraints at this level have been considered as well.



However, often relaxed versions of these constraints are needed, and this increases considerably the complexity of computing value-aligned policies. To obtain efficient algorithms that make value-aligned decisions considering admissibility relaxation, we propose the use of learning techniques, in particular, we have used constrained reinforcement learning algorithms. In this paper, we present two algorithms, \mbox{$\epsilon$-ADQL} for strategies based on local alignment and its extension \mbox{$\epsilon$-CADQL} for a sequence of decisions. We have validated their efficiency in a water distribution problem in a drought scenario.


\end{abstract}
\end{frontmatter}

%


\section{Introduction}

Integrating human values into practical reasoning is a problem that has been considered only recently in computer science and autonomous systems \cite{weide2010practical,bench2012using}. These and other attempts to incorporate human values into the reasoning and decision-making schemes of intelligent software agents can be labelled into the emergent field of \emph{value awareness engineering}~\cite{DBLP:journals/corr/abs-2302-08759}. Proposals for modelling value-based decision processes of autonomous agents are often based on preferences over states or actions~\cite{montes2022synthesis,leraleri2022aggregation}, which are then extended to sequential decisions. Other approaches~\cite{christiano2023deeprlpreferences,furnkranz2012preference} set out from observed sequences of actions (plans) and then learn preferences over states or actions through (inverse) reinforcement learning~\cite{ng2000algorithms}.

In \cite{andres2023eumas} we have argued that there is a need to look into value-alignment from a path-level perspective in a careful way. In particular, without denying the usefulness of using aggregations of state or action preferences to measure the alignment of sequential decisions (paths), we have suggested that certain sequences of actions should not be admissible to a value-aware agent, even though they show a good overall aggregated value-alignment~\cite{andres2023eumas}. This is captured through the notion of \textit{value-admissible behaviours}, i.e. criteria defining minimally aligned sequences of decisions, which control more precisely \textit{how} a value-aligned path should be.


In this paper, we look into the problem of learning decision policies that represent admissible behaviours that are highly aligned with a certain value. In order to solve this problem, reinforcement learning (RL)~\cite{sutton2018reinforcement} and, in particular, constrained versions of it constitute state-of-the-art tools. However for infinite horizon processes, where a reward should be maintained over time instead of reaching a goal, average reward MDP's (Chapter 10, Sutton and Barto~\cite{sutton2018reinforcement}) offer a better approach than the usual discounted setup.

Based on the above insights, in this paper we propose new RL methods called $\epsilon$-local (Constrained) Average
Double Q-Learning ($\epsilon$-ADQL and $\epsilon$-CADQL, respectively), implemented by borrowing concepts from constraint RL: RCPO (Reward Constraint Policy Optimization)~\cite{RCPOtessler2018reward} and CDQL (Constrained Deep Q-Learning)~\cite{deepConstrainedQLearning2020kalweit}. We test them in a continuous water distribution procedure in a simulated environment where rewards model directly a numerical representation of value-alignment at state-level. Specifically, $\epsilon$-ADQL will find a policy maximizing expected mean future reward.
In addition, the constrained version $\epsilon$-CADQL aims at minimising potential \textit{violations} of state-level admissibility criteria, intending to avoid as much as possible states whose alignment with a certain value (modelled by the semantics function) drops below a threshold $\tau$.

The paper is structured as follows. Section~\ref{sec:related-work} discusses related work regarding value awareness and reinforcement learning. In Section~\ref{sec:behaviours}, we outline our world model and introduce different notions of value-admissible behaviours in the context of learning problems. In Section~\ref{sec:use-case}, we motivate and explain our use case on water distribution considering the value of equity. In Section~\ref{sec:learning} we model the problem of learning value-aligned admissible paths in our use-case and provide details on the $\epsilon$-ADQL and $\epsilon$-CADQL RL algorithms. In Section~\ref{sec:evaluation} we test the algorithms in simulations and compare them to a baseline local policy. Finally, in Section~\ref{sec:conclusions} we present our conclusions and future work.

\section{Related work}\label{sec:related-work}

The value-alignment problem has been formalized in various ways for decision-making. One of the first communities introducing values in their algorithms was practical reasoning~\cite{weide2010practical} and value-based argumentation~\cite{bench2012using}. Computational representations of values are key to the recent field of \textit{value-awareness engineering}. 

EU ethics guidelines on trustworthy AI\footnote{digital-strategy.ec.europa.eu/en/library/ethics-guidelines-trustworthy-ai} mention a series of values including privacy, fairness, explainablity, non-maleficence etc. that must be respected during all stages of the design and development of an AI system to ensure its responsible deployment. Schwartz puts forward a taxonomy of universal values that transcend specific actions or situations in a system~\cite{schwartz2012overview}. However, for practical reasoning such values need to be grounded in a particular context~\cite{DBLP:journals/corr/abs-2302-08759}.


Works borrowing from the consequentialist tradition of computational ethics usually model values in terms of preferences over states or actions~\cite{values2021nardineold,bench2012using,montes2022synthesis}. Montes \textit{et al.}~\cite{montes2022synthesis} use \emph{semantics functions} to define the value-alignment of states. This information  is then extended to sequential decisions via aggregation functions in order to find norms that can regulate a multi-agent system's behaviour in a value-aware fashion. However, they do not examine in detail how far the resulting evolution of the system (i.e. joint paths traversed by agents) really certifies an acceptable alignment at path-level. 
Lera Leri \textit{et al.}~\cite{leraleri2022aggregation} define value-alignment based on actions that promote or demote values. They use that formulation for the task of \textit{value system} aggregation (i.e. ranking decisions taking perspectives of different values into account) but do not study sequences of actions. 
The notion of value-admissible behaviour put forward in~\cite{andres2023eumas} introduces constraints on value-alignment at path-level. However, the work does not provide effective algorithms to determine admissible sequential decisions with a suitable level of value-alignment.

Regarding applications of machine learning to value-alignment, inverse reinforcement learning~\cite{ng2000algorithms} has been used to infer values (preferences over states or actions), setting out from observed sequences of actions~\cite{christiano2023deeprlpreferences,furnkranz2012preference}. The problem of determining optimally aligned sequences of actions given a certain value is often modelled as a mathematical optimization problem~\cite{sudipto2012anonymizing}. 
In the context of this paper, we are particularly interested in RL techniques capable of generating suitable value-aligned sequences of actions that respect certain admissibility constraints. To this respect, it is important to highlight the multi-objective RL approach developed by Rodríguez-Soto \textit{et al.}~\cite{manel2022ethical}, integrated into a norm-abiding and value-aligned (i.e. \textit{ethical}) decision-making learning environment, that uses specific ethical rewards and RL values. Still, this approach cannot express path admissibility criteria that are not reward-shapable. 

In the field of constrained RL~\cite{altman1999constrained}, Tessler \textit{et al.}~\cite{RCPOtessler2018reward} propose an algorithm (RCPO) to deal with state-action constraints (formulated as inequalities) learning the optimal Lagrange multiplier to shape the rewards optimally. Dalal \textit{et al.}~\cite{dalal2018safe} introduce the concept of \textit{safety} (in critical environments) directly adding to the policy a safety layer that analytically solves an action correction formulation per each state, \textit{predicting} constraints. Both approaches are able to deal with certain behaviours but for a discounted setup, which does not fit well with long-term continuous problems~\cite{sutton2018reinforcement} that we are particularly interested in. Another solution, Constrained Q-learning, by Kalweit \textit{et al.}~\cite{deepConstrainedQLearning2020kalweit}, enforces hard constraint satisfaction during and after the training process, though its efficiency is limited by the behaviours' complexity. 






\section{Value-admissible behaviours}\label{sec:behaviours}

As a first step towards agents learning value-aligned action policies, in this section, we sketch the world model we used and introduce different types of value-admissible behaviours that a value-aware agent may want to choose from.

Setting out from~\cite{montes2022synthesis}, we assume that an agent's decision-making component represents the world as a labelled transition system, called \textbf{decision world}.

\begin{definition}[Decision world]
    A decision world is a triplet $(\mathcal{S},\A,\T)$ with the following elements.

\begin{itemize}
    \item \textbf{States} $\mathcal{S}$, representing the MAS completely in each situation.
    \item \textbf{Actions} $\A$, representing the MAS joint actions or decisions.
    \item \textbf{Transitions} $\T \subset \mathcal{S} \times \A \times \mathcal{S}$, representing available actions connecting each pair of states. Denoted with $s \xrightarrow{a} s'$, where $s, s' \in \mathcal{S}$ and $a \in \A$. We also define $\A(s)$ as the actions accessible in the system from state $s$ (i.e. the set of actions $a\in\A$ such that exists some $s\xrightarrow{a}s' \in \T$).

    \item \textbf{Paths} $\mathcal{P}$, representing joint transitions (sequences of decisions), e.g. a path of length $n$ from $s_0$ to $s_n$ would be represented as: $P = s_0 \xrightarrow{a_1} s_1 \xrightarrow{a_2} \dots \xrightarrow{a_n} s_n$.
\end{itemize}
\end{definition}
While in~\cite{montes2022synthesis} a notion of final or \textit{goal} states for the system to reach via paths where considered, in this paper goals are considered implicit and policies can imply infinite sequences of actions.


Following Weide et al.~\cite{weide2010practical} or Sierra et al.~\cite{values2021nardineold}, we assume a value preference among states based on a preorder relation $\sqsubseteq_v$, which we call \textbf{perspective} or \textbf{value preorder}, i.e. given $s$ and $s'$, two states, $s \sqsubseteq_v s'$ means that $s'$ is at least as preferred as $s$ w.r.t.\ the value $v$.

However, to simplify the computational representation~\cite{andres2023eumas}, in this paper we will quantify the above relationships using \textit{unbounded}\textit{semantics functions} \cite{montes2022synthesis}:\footnote{Original definition from Montes and Sierra uses $[-1,1]$-bounded functions, which is used to model both promotion and demotion of the value. For this theory, those specific bounds are not mandatory.}
\begin{definition}[(Unbounded) Semantics function]\label{def:semanticfunctionMS}
  The alignment of state $s$ with a value $v$ is described by an \textit{unbounded}
  \textbf{semantics function}  $ f_v : \mathcal{S} \longrightarrow \R $, where $f_v$ is directly proportional to the promotion of $v$.
\end{definition}


Our objective is to find a policy which leads to a path that is well aligned with a certain value. Specifically, we want to maximize the values of the semantics function along possibly an infinite sequence of states, using a certain \textit{aggregation function}~\cite{montes2022synthesis} as a metric:

\begin{definition}[value-alignment of a path]\label{def:agg} Given an aggregation function $agg$, and a semantics function $f_v$, we define the \textbf{value-alignment of a path} $P= s_0 \xrightarrow{a_1} \dots \xrightarrow{a_n} s_n$ (or its aggregated alignment) as:
$$agg_v(P) = agg(\{f_v(s_0),\dots,f_v(s_n)\})$$

\end{definition}

As we are considering an infinite horizon, we opt to use the \textit{average} as the aggregation function (thus making path length irrelevant).

However, in certain circumstances a path with maximum average aggregated alignment may not be \textit{admissible} to a value-aware decision-maker, e.g. due to an extreme variability of the value-alignment of the states it traverses. For instance, in a water distribution scenario, all assignments that at some point in time leave stakeholders without a minimum amount of water necessary for basic needs should not be considered, even though ``on average'' they achieve an equitable water distribution. Setting out from~\cite{andres2023eumas}, in the following we present a redefinition of the concept of \textit{value-admissible behaviours} based on semantics functions.

\begin{definition}[Value-Admissible Behaviour]
    A \textbf{value-admissible} \textbf{behaviour} for a value $v$ is a constraint criterion for plans $\mathcal{P}$ that characterizes the subset $B(\mathcal{P}, f_v)$ that are admissibly aligned with the value, based on state/action-level semantics $f_v$. 
\end{definition}

In real-world draught scenarios, legal requirements establish that the equity of water distribution has to be assured at all times~\cite{andres2023eumas}. This is expressed by the following \textit{local} behaviour in terms of the values of a semantics function.

\begin{definition}[Local behaviour]
    Given a set of paths $\mathcal{P}$ and a semantics function $f_v$, the local behaviour, $B_{local}$ is defined as the set of paths built by maximizing the value function at each step: 
    \begin{align*}
        B_{local}(\mathcal{P},f_v) &= \{P \in \mathcal{P}\ |\ \forall s \xrightarrow{a}t  \in P:\\&f_v(t) = \max \{f_v(t')\ |\ \exists a' \in \A(s): s\xrightarrow{a'}t' \}\}
    \end{align*}
\end{definition}
Still, maximizing the overall aggregated equity (i.e. the alignment of paths with the value of equity) is of high importance. In special cases, both characteristics go hand in hand. For instance, in~\cite{andres2023eumas} equity semantics functions adhering to the Pigou-Dalton principle~\cite{moulin2004fair} provide that the locally admissible paths are also those with the highest aggregate value-alignment. However, with general semantics functions and in more complex environments this need not be the case.

Therefore we introduce the following relaxed notions of the aforementioned local behaviour that will constitute the basis of our learning approach put forward in Section~\ref{sec:learning}.



\begin{definition}[$\epsilon$-local behaviour]\label{def:epsilonstrat}
    Given a set of paths $\mathcal{P}$, $\epsilon > 0$ and a semantics function $f_v$, the \textbf{$\epsilon$-local behaviour}, $B_{\epsilon}$ is defined as: 
    \begin{align*}
        B_{\epsilon}(\mathcal{P},f_v) &= \{P \in \mathcal{P}\ | \ \forall s \xrightarrow{a}t  \in P:
        \\&f_v(t) \geq \max \{f_v(t')\ |\ \exists a' \in \A(s): s\xrightarrow{a'}t' \} - \epsilon\} 
    \end{align*}
\end{definition}
The epsilon-local behaviour allows traversing new paths, by slightly relaxing the strict equity-preserving aspects of the local one. We expect that this relaxation of immediate equity prosecution will lead, with a certain $\epsilon$-\textit{local policy} (i.e. $\epsilon$-local behaviour compliant) and a sufficiently big $\epsilon$, to paths with much higher aggregation, traversed with fairly legally justifiable actions.
Given a state $s$ and a value $\epsilon$, we denote by $\A_{\epsilon}(s)$ the actions that would be admissible to be executed in state $s$ in an $\epsilon$-\textit{local policy}.

Apart from the $\epsilon$-local behaviour, to guide our policies into traversing globally admissible states (i.e. the semantics of the value for every state traversed is above some threshold $\tau>0$), we propose the $\tau$-constrained behaviour:


\begin{definition}[$\tau$-constrained behaviour]\label{def:tauconstrained}
    Given a set of paths $\mathcal{P}$, $\tau > 0$ and a semantics function $f_v$, the \textbf{$\tau$-constrained behaviour}, $B_{\tau}$ is defined as:
    \begin{align*}
            B_{\tau}(\mathcal{P},f_v) &= \{P \in \mathcal{P}\ | \ \forall s \xrightarrow{a}t  \in P:
        f_v(t) \geq \tau \} 
    \end{align*}
\end{definition}
In short, a $\tau$-constrained policy would lead to paths where the alignment of all states in the path with respect to a semantic function $f_v$ has a lower bound on a threshold $\tau$.


\section{Use Case}\label{sec:use-case}

\begin{figure}
    \centering
    \includegraphics[trim=0cm 0cm 0cm 0cm,clip,width=0.7\columnwidth]{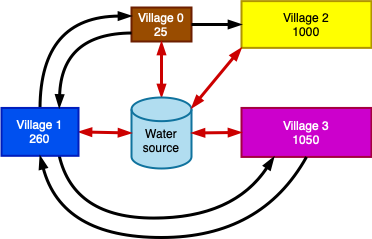}
    \caption{The water distribution problem schema (Red arcs for two-way roads, directed black arcs for one-way ones)}
    \label{fig:villages}
\end{figure}

In this paper, we consider a use case around water distribution where equity (or fair distribution) is the value to be preserved. This field has been widely analyzed using socio-cognitive agents \cite{antoniperello2021water} but here we use a simpler, yet just complex enough, scenario that is sufficient for illustrating the proposed concepts. The tasks consist of distributing water from a reservoir (source of water) to 4 villages with different populations and that are connected through a road network.
Figure~\ref{fig:villages} represents the map of the problem. 

 A tanker truck able to traverse these roads is in charge of distributing the water. 
 We assume that this vehicle has a limited  capacity $C$ (here, $C = 60000l$) and the process continues until a fixed amount of water $T$ has been distributed (i.e. the available water resources). 

On each step, the vehicle can take an action which consists of moving from its current \textit{position} (i.e. either a village or the reservoir) to another point and discharging some water (at villages) or refilling the tank (at the reservoir). We encoded some simple rules regarding possible movements for basic delivery and time efficiency:

\begin{itemize}
    \item If the vehicle visits a village that has no outgoing connection to other villages (e.g. Village $2$ in Figure~\ref{fig:villages}), it will dispense all its current amount of water.
    \item Unless the vehicle is empty, it cannot return back to refill at the reservoir.
    \item Each time the vehicle returns to the source, it fills up completely.
\end{itemize}

For simplicity, we assume the truck takes equal time (approx. one hour) to reach any node (village/source) reachable from any immediately adjacent node (connected by a road). 

In our scenario, we consider that each village consumes the water based on different consumption patterns, as described in the next subsection.

\subsection{Estimation of water consumption}\label{sec:legal}
According to the World Health Organization (WHO), between 50 and 100 litres of water per person per day are necessary to ensure that the most basic needs are met and few health concerns arise. However, average water use is 200 to 300 litres per person per day in most countries in Europe.\footnote{Resolution 64/292, 07/28/2010, of the United Nations General Assembly, vid. \url{https://www.un.org/spanish/waterforlifedecade/human\_right\_to\_water.shtml}.}

In Spain, the main use of water is for irrigation and agricultural use, which accounts for approximately 80.5\% of this demand, followed by urban supply, which represents 15.5\%. The remainder is for industrial use~\cite{perte2022gob}. Of all the water uses, the priority is urban water supply.\footnote{Royal Decree 1/2001, of July 20, approving the Revised Text of the Water Law, Article 60.} The regulations have established that the net or average consumption endowment, as a minimum objective, must be at least 100 litres per inhabitant per day.\footnote{Royal Decree 3/2023, of January 10, establishing the technical-sanitary criteria for the quality of drinking water, its control, and supply, Article 9.}

This information inspires the idea to model that different societies (or villages in our case) would have different water consumption patterns. In our experiments, we will model different  consumption rates per \textit{time step} (the period of one movement of the tanker vehicle). We consider that the consumption will increase as water availability in a village increases (e.g., due to a diversification of uses), but differently depending on each village:

\begin{itemize}
    \item Village 0 (Brown-$25$ inhabitants): This village consumes $4l$ per inhabitant per hour in scarcity, unless the available water is above $350l$ per inhabitant, where it will consume $100l$ per inhabitant per hour due to large irrigation needs.
    \item Village 1 (Blue-$260$ inhabitants): In this village only $3.5l$ per inhabitant per hour is used unless there is more than $250l$ per inhabitant. Then, $9l$ per inhabitant per hour will be consumed.
    \item Village 2 (Yellow-$1000$ inhabitants) In this case  $3.5l$ is consumed per inhabitant per hour if less than $350l$ per inhabitant is in reserve. Otherwise, per inhabitant consumption rises to $50l$ per hour.
    \item Village 3 (Magenta-$1050$ inhabitants). In this village, the consumption rate will be $3.5l$ up to a reserve of $100l$ per inhabitant. After that, the consumption will be $16l$ per inhabitant.
\end{itemize}

\subsection{Modeling states, transitions, and paths}

A possible state- and transition-based modelling of the decision problem for water distribution among the four villages is as follows:

- \textbf{States}: conceptually, we will represent a state $s\in \mathcal{S}$ as a list $(x_0, x_1, x_2, x_3, p, c)$ where each value $x_j \in \R_{\geq0}$ represents the amount of water per inhabitant in village $j$; $p \in \{-1, 0,1,2,3\}$ represents the current position of the vehicle ($-1$ for source); and $c$ stands for the current amount of water in the vehicle. 

- \textbf{Actions}: an action $a \in \A$ is a pair $(p, d)$ indicating the place $p \in \{-1,0,1,2,3\}$ to go to by the vehicle and $d \in (0, C]$ the water to dispense after arriving at $p$ (irrelevant when $p = -1$, as then the vehicle will just refill). As described before, the available actions depend on each state.

- \textbf{Transitions}: a transition from state $s = (x_0, \dots, x_3, p, c)$ to state $s' = (x_0', \dots, x_3', p', c')$ by applying action $a=(p',d)$ means that $s'$ is calculated by setting $c' = c-d$ if $p \neq -1$ (else, $c'=C$ for refilling) and calculating $x_i'$ by first letting the vehicle move (and dispense the water to $p$) and then applying the corresponding water consumption rate to each village as described in Section \ref{sec:legal}.

The decision-making solution in the distribution of water between villages is a (deterministic) Markov policy. We are interested in finding policies that generate certain behaviours (e.g., fulfil certain admissible criteria) and also lead to paths with high value-alignment with respect to a semantics function and aggregation function.

\subsection{Modelling equity}\label{sec:principio-de-equidad}\label{sec:equality}


In the water distribution scenario, we are interested in the value of equity. Thus, we need to define what it means to have a fair or equitable distribution of water in our system at each moment (i.e. how much our states of the system are aligned to the equity value). We will use a semantics function that reflects equity in terms of water availability per person, not per village, assuming each person has equal access to the water of the village he or she belongs to. 
Specifically, let $n_i$ denote the population of village $i$ and $s = (x_0, \dots, x_3, p, d)$ be a state of the environment. We define the \textit{water distribution} of state $s$, $D(s)$ by:
$$D(s) = (\underbrace{x_0, \dots, x_0}_{n_0 \text{ times}},\dots, \underbrace{x_3,\dots,x_3}_{n_3\text{ times}})$$

In this distribution, each person counts with his or her own water available water resources. For example, if village 0 (with 25 inhabitants) has $20l$ per inhabitant, there are 25 entries with $20l$.

Based on $D$ we define the semantics function $f_{gini}(s)=1-GI(D(s))$, where $GI$ is the Gini Index, a widely accepted inequality metric in the literature \cite{montes2022synthesis,plata2015elementary,Guo2023Socialdilemma}. Note $f_{gini}$ is bounded in $[0,1]$, and is proportional to equity. 

In the next section, we will use the $3$ behaviours presented in Section~\ref{sec:behaviours} for assessing equitable paths. We will define settings for learning policies that maximize the aggregated alignment of paths and also implement $\epsilon$-local / $\epsilon$-local and $\tau$-constrained behaviours. We will compare those policies with a policy that simply adheres to a local behaviour. 

\section{Learning behaviour-compliant policies}\label{sec:learning}

In our scenario, given a current state $s$, we can predict the value-alignment of the resulting states if we would apply any possible action to $s$. This can be done by simply calculating $f_{gini}$ on the estimated resulting states. Thus, a policy adhering to a local behaviour can be implemented easily by seeking the best action among all possibilities in each moment.
However, using brute force search to find a policy that maximizes the value-alignment of an execution path is often not feasible, even for simple tasks. In our approach, we propose to use Reinforcement Learning (RL) \cite{sutton2018reinforcement} as a technique to learn such policies.

Our use case (Section~\ref{sec:use-case}) can be represented by a Markov Decision Process (MDP) where in any state a more or less \textit{equity-valuable} action can be taken. This equity-valuable concept is directly represented via state-action \textit{rewards} $r(s,a)$, which are then used to define the RL \textit{value} $V(s,a)$ representing future expected reward (i.e. ``long-term'' equity) in the system. This value function $V(s,a)$ is used by the policy to select action $a$ from $s$. Specifically, for any state $s_j$ at a step $j$, we calculate the reward of applying an action $a_j$ from that state such that $s_j \xrightarrow{a_j} s_{j+1}$ as $r_j(s_j,a_j) = f_{gini}(s_{j+1})$. 

Given this reward definition, our aim is to learn policies that maximize path value-alignment and that are at the same time \textit{behaviour-compliant} with $\epsilon$-local and/or $\tau$-constrained behaviours.

In order to keep the learning problem tractable,  we apply a simplification to both state and action spaces.  
With regard to actions, we only allow deliveries of water (the value $d$ from action pair $(p,d)$) which are multiples of a fixed quantity. In our experiments, we used $15000l$ and allowed only deliveries of $0l$, $15000l$, $30000l$, $45000l$ and $60000l$ (which is the capacity of the truck). 

With regard to states, we reduce the state space by converting the values of water per inhabitant into intervals, called \textit{levels}\footnote{To calculate $f_{gini}$ we still use the original states.}. We use two parameters to define these levels: i) $m$ --- the minimum legal water requirement for a person per day (in our case $100l$), and ii) $M$ --- the desired water level per person in all villages, e.g., the level at which no further limitations are applied (in our case $350l$). Using $m$ and $M$ we define the following levels:
\begin{itemize}
    \item Level $0$ (\textit{red} level), representing water levels below $m - (M+m)/2$ (or $0$ as a lower bound)
    \item Level $L$ (\textit{green} level), representing levels over $M + (M+m)/2$. 
    \item Levels between $0$ and $L$, representing evenly distributed intervals between water levels $0$ and $L$. The number of \textit{hidden levels} is $H$, a hyperparameter of the policy.
\end{itemize}

In our approach, we apply Double $Q$-learning~\cite{hasselt2010doubleDQL}, in its basic tabular version as the basic RL algorithm.\footnote{Double $Q$-learning reduces training biases of normal $Q$-learning (though reducing sample efficiency). This RL algorithm choice is not critical, though, as our proposed Algorithms~\ref{alg:epsilon} and \ref{alg:epsilon-constrained} will work fine with typical $Q$-learning too.} Moreover, we change the typical RL expected discounted reward maximization goal to an average reward maximization goal equivalent to our path aggregated alignment notion (Chapter 10, Sutton and Barto~\cite{sutton2018reinforcement}).
This change is motivated also by the fact that we need to cope with a continuous goal problem. 

\subsection{Learning policies for $\epsilon$-local behaviours}
We first present Algorithm~\ref{alg:epsilon} ($\epsilon$-ADQL, $\epsilon$-local Average Double Q-learning) for learning $\epsilon$-local behaviour adherence policies (i.e. $\epsilon$-local policies) while maximizing average state value-alignment. It is directly based on CDQL by Kalweit \textit{et al.}~\cite{deepConstrainedQLearning2020kalweit}. The safe set $S_C(s)$ defined by \cite{deepConstrainedQLearning2020kalweit}, corresponds in our case to the set of $\epsilon$-admissible actions $\A_\epsilon(s)$.

Algorithm~\ref{alg:epsilon} is able to learn a policy $\pi^*$ which finds the path with optimal average alignment of the path states while adhering to an $\epsilon$-local behaviour. The learned policy samples paths with the following formula:
\begin{equation}\label{eq:policy_sample}
    \pi^*(s) = \argmax_{a\in \A_\epsilon(s)}{Q(s,a) + Q'(s,a)}
\end{equation}

\begin{algorithm}[H]
\caption{Epsilon-local Average Double Q-Learning $\epsilon$-ADQL}\label{alg:epsilon}
\begin{algorithmic}[1]
\State{Initialize two $Q$-tables \cite{hasselt2010doubleDQL} $Q$ and $Q'$, exploration rate $p\in(0,1)$} and average reward estimator $\hat{r}=0$ (and its learning rate $\beta \in(0,1)$).
    \For{optimization step $o=1,2,\dots, N$}
    \State $j=0$
    \State $s_0\gets$  \textproc{Reset}(environment)
    \While {environment is not done}:
    \State Sample $n \sim \textit{Uniform}([0,1])$
    \If {$n < p$}{ choose $a_j \in \A(s_j)$ randomly}
    \Else 
    \State Compute $\A_\epsilon(s_{j})$ ($\epsilon$-local valid actions)
    \State{$a_j \gets \argmax_{a\in \A_\epsilon(s_j)} Q(s_{j},a)+Q'(s_{j},a)$}\footnotemark\Statex\Comment{Equation~\ref{eq:policy_sample}}
    \EndIf
    \State $(s_{j+1},r_j) \gets $ \textproc{Step}(environment, $a_j$)\Statex\Comment{Recall $r_j = f_{gini}(s_{j+1})$}
    \State $Q_1, Q_2 $= \textproc{PermuteRandomly}(Q,Q') \Statex\Comment{(Changes to $Q_i$ are kept back to $Q$ and $Q'$)}
    \State Compute $\A_\epsilon(s_{j+1})$
    \State $a_{Q_2,j+1} = \argmax_{a\in \A_\epsilon(s_{j+1})} Q_2(s_{j+1},a)$
    \State $\delta_j \gets r - \hat{r} + Q_2(s_{j+1}, a_{Q_2,j+1}) - Q_1(s_j, a_j)$
    \Comment{TD er.}
    \State $\hat{r} \gets \hat{r} + \beta\delta_j$\Comment{Average reward estimation \cite{sutton2018reinforcement}}
    \State $Q_1(s_j, a_j) \gets \alpha\delta_j$\Comment{Update $Q_1$ with $Q_2$ criterion.}
    \State $j \gets j + 1$
    \EndWhile
    \State (Optional) Decrement $p$ w.r.t. $o$.
    \EndFor
\end{algorithmic}

\end{algorithm}

\footnotetext{With some abuse of notation, $Q(s,a)$ denotes in reality the $Q$-table value of the leveled representation of state $s$ under action $a$.}

\subsection{Learning policies for $\tau$-constrained and $\epsilon$-local behaviours}

Algorithm~\ref{alg:epsilon-constrained} presents an algorithm for learning policies that adhere to both: $\tau$-constrained and $\epsilon$-local behaviours. We call this algorithm $\epsilon$-CADQL ($\epsilon$-local Constrained Average Double Q-Learning).


We should be aware that we cannot just proceed as with the $\epsilon$-local case, because there might be situations where the semantics function of all possible direct future states will be below $\tau$. This could happen, for example, when initializing the world to a state $s$ with an alignment value much lower than $\tau$. In such cases, we would like the algorithm to measure/count the $\tau$ bound \textit{violations} of sampled paths for learning policies minimizing the proportion of violations per path length (\textit{violation ratio}, $V$). To do so, we modified our previous Algorithm~\ref{alg:epsilon}, adding an implementation of an intelligent reward shaping approach via Lagrangian multipliers as proposed in \cite{RCPOtessler2018reward}. Following their method, RCPO, we first adopt the following \textit{penalties} $c(s,a)$ and corresponding \textit{constraint} $C(s_0)$ for a path $s_0 \xrightarrow{a_0} \cdots \xrightarrow{a_{M-1}}s_M$:

\begin{align}\label{eq:penalty}
    c(s,a) &= \begin{cases}
{1}, &\text{if violation detected applying $a$} \\
{0}, &\text{otherwise}
\end{cases}\\
C(s_0) &= \sum_{j=0}^M c(s_j,a_j) \leq 0
\end{align}

This definition ensures local minima of the constraint are feasible solutions, and $C(s_0) > 0$ represents the total behaviour violations of a path.
As done in \cite{RCPOtessler2018reward}, we use a learned Lagrangian parameter $\lambda$, and we also model a suitable $\Gamma_\lambda$ projection, as follows.
$$\Gamma_\lambda (x) = \min\{x, \bar{R}/\bar{V} \}$$
Here, $\bar{R}$ is the expected path average reward (without any penalties considered) and $\bar{V}$ is the expected \textit{violation ratio}. 
Simply put, all the elements above serve to work with a modified aggregated reward given by calculating in each transition $s_j\xrightarrow{a_j}s_{j+1}$ (with observed reward  $r_j$) a modified reward by subtracting to $r_j$ a penalty of $c(s_j,a_j)$ weighted by the factor $\lambda$ (which approximates to $\bar{R}/\bar{V}$ as Algorithm~\ref{alg:epsilon-constrained} evolves), and then averaging over all transitions.
With this process, one can see that if $\lambda \approx \bar{R}/\bar{V}$, the algorithm will assign a (modified) aggregated reward value of $0$ to the paths that get exactly a \textit{violation ratio} of $\bar{V}$; a negative value to those with a bigger ratio; and a positive value to paths with a smaller one. In particular, it is guaranteed that non-violating paths are preferred over violating paths, no matter their respective values of their value-alignment.

\begin{algorithm}[ht]
\caption{Epsilon-local Constrained Average Double Q-Learning}\label{alg:epsilon-constrained}
\begin{algorithmic}[1]
\State{Initialize $Q$, $Q'$, $j=0$, exp. rate $p\in(0,1)$, $\hat{r}, \hat{V}, \hat{R}=0$ (and $\beta \in (0,1)$), penalty factor $\lambda > 0$},  and rates $\alpha > \beta_R > \beta_V > \alpha_\lambda$ (following \cite{RCPOtessler2018reward}).
    \For{optimization step $o=1,2,\dots,N$}
    \State $s_0\gets$ \textproc{reset}(environment), $j \gets 0$, $R_o \gets 0$, $C_o \gets 0$
    \While {environment is not done}:
    \State $(s_{j},a_j,s_{j+1},r_j) \gets $ Algorithm~\ref{alg:epsilon}, lines 6-12.
    \State $R_o \gets R_o + r_j$
    \If{$c(s_j, a_j) > 0$}
    
    \State $r_j \gets r_j - \lambda_o c(s_{j},a_j)$ 
    \State $C_o \gets C_o + 1$

    \EndIf
    
    \State $Q_1, Q_2 $= \textproc{permuteRandomly}(Q,Q')
    \State Update $Q_1$ and $Q_2$ as for Algorithm~\ref{alg:epsilon}, Lines 14-16.
    
    \State $j \gets j + 1$
    
    \EndWhile
    \State (Optional) Decrement $p$ w.r.t. $o$.
    \State $\lambda_{o+1} \gets \hat{\Gamma}_\lambda(\lambda_{o} + \alpha_\lambda \cdot C_o)$ \Comment{RCPO Lagrange update \cite{RCPOtessler2018reward}}
    \State $\hat{V} \gets \beta_V \frac{C_o}{j} +  (1 - \beta_V)\hat{V} $
    \State $\hat{R} \gets \beta_R \frac{R_o}{j} +  (1 - \beta_R)\hat{R} $

    \EndFor
\end{algorithmic}

\end{algorithm}

At the beginning of the algorithm, $\lambda$ is set to $0$.  This encourages exploration even while committing infractions. As the training process evolves, the algorithm will learn better paths allowing $\lambda$ to grow in the process (diminishing the possibility of incurring in violations, when possible). Though this might create bias, we update $\lambda$ much slower than the policy (convergence idea from \cite{RCPOtessler2018reward}), thus, temporal biases are less meaningful. After each completed episode 
we approximate $\bar{R} \approx \hat{R}$ and $\bar{V} \approx \hat{V}$ on the go with very conservative exponential weighted averages $\beta_V,\beta_R \approx 0.001$, after calculating the next $\lambda$ using approximate projection $\hat{\Gamma}_\lambda (\lambda) = \min\{\lambda, \frac{\hat{R}}{\hat{V}} \}$.

After learning, the result of Algorithm~\ref{alg:epsilon-constrained} is a policy (defined through equation \ref{eq:policy_sample}) which implements an $\epsilon$-local and also a $\tau$-constrained behaviour. 

\section{Evaluation}\label{sec:evaluation}
The objective of the evaluation is: i) to see the advantages of relaxing the local policy (using the $\epsilon$-local concept) towards maximizing the value-alignment of paths, and ii) to see implications of the simultaneously $\tau$-constrained and $\epsilon$-local policy trained via Algorithm~\ref{alg:epsilon-constrained}. 

\subsection{Training setup}\label{sec:hyperparameters}\label{sec:training}
The training environment was programmed using the former OpenAI Gym \cite{gym2016openai} library, now held by \cite{gymnasium}. 
The environment was extended to add a method to get the sets $\A_\epsilon(s_j)$ after every step. In the experiments, we used the connected map in Figure~\ref{fig:villages}, with village parameters from Section~\ref{sec:use-case} and the following hyperparameters:
\begin{itemize}
    \item $\alpha = 0.03$, $\alpha_\lambda = 0.0003$; $\beta = 0.01, \beta_V=\beta_R= 0.001$.
    \item $H = 5$ \textit{hidden levels} for distribution representation.
    \item $\epsilon = 0.1$. $\tau = 0.7$.
    \item $N=30000$ iterations. Exploration $p=0.3 \to 0$, as $o \to N$.
\end{itemize}

In the training processes, an episode finishes when $1440000$ litres of water are distributed. Furthermore, in each episode, the initial state is randomly \textproc{Reset} with village water levels sampled from a uniform distribution between $0$ and $600l$.

After the training processes, the obtained policies have been applied in an evaluation scenario. It has a default initial state of $(0, 300, 200, 200, -1, 60000)$ and the experiment is run until $3000000l$ of water are distributed.

\subsection{Experiments}

\begin{figure}
    \centering
    \includegraphics[width=0.5\columnwidth,height=5.5cm]{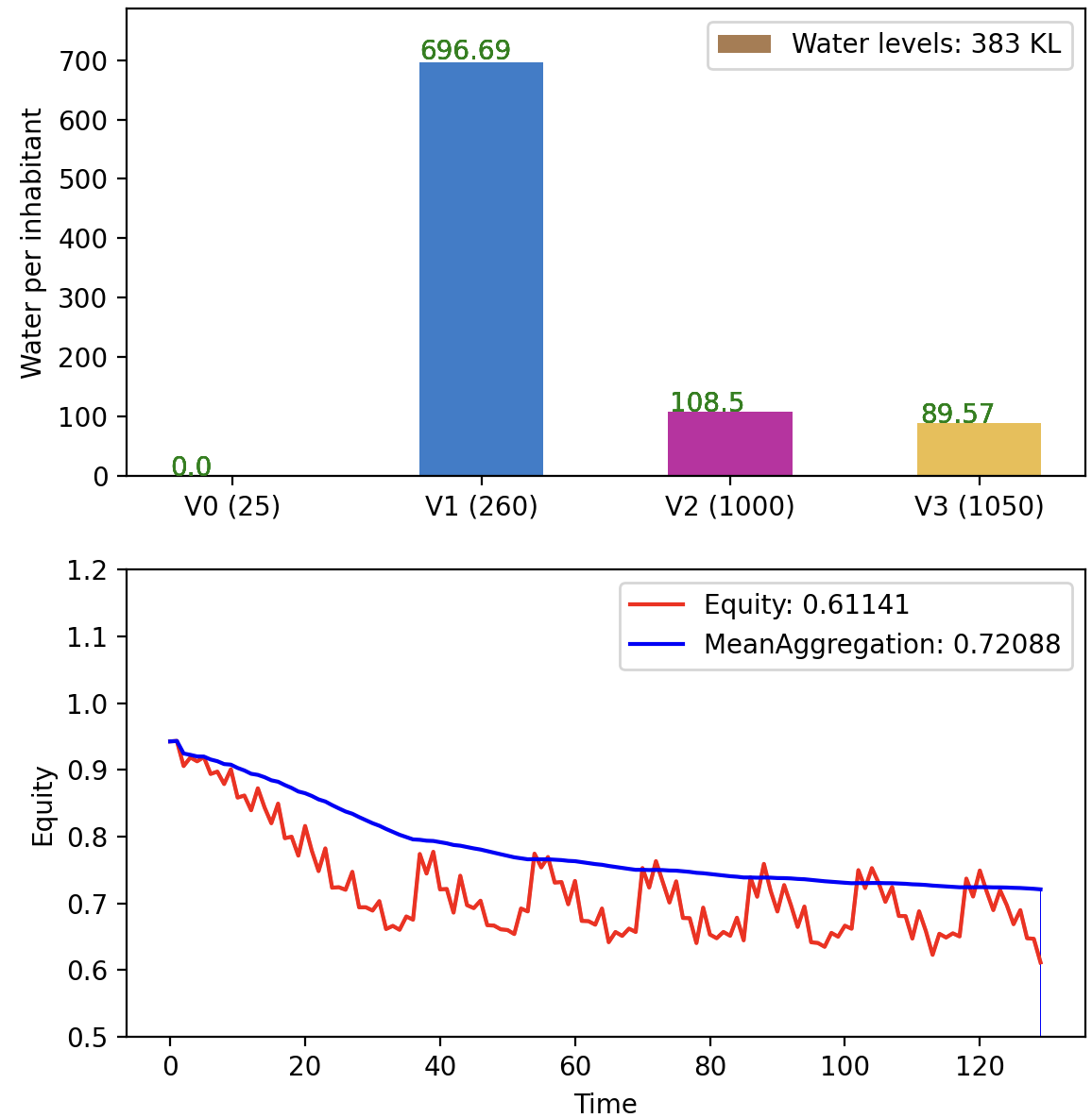}
    \vspace{-0.5cm}
    \caption{Local policy. Top: distribution of water and the total amount of available water at the end of the episode, Bottom: Equity (rewards) at each step and aggregated average (i.e. average reward until each step).}
    \label{fig:local_policy}
\end{figure}

First, we want to highlight the advantages that can be obtained by using the policy that implements an $\epsilon$-local behaviour over following the simple local strategy. The performance of the \textit{local policy} is shown in Figure~\ref{fig:local_policy}. Notice, that apart from making the equity criteria worse over time, Village $1$ gets too much water. This is possibly due to the fact that village 1 has many connections and thus, the truck is more often forced to dispose of water to that village.

Analysing the results over time, it is noticeable that the truck never visits Village $0$ (with just 25 inhabitants). This is because from the point of view of equity, in the short term, it is normally better to bring water to populous villages. Furthermore, the truck would discharge too much water for this small village, massively increasing its water per inhabitant, leading to less equity. The local policy is not able to take into account the water consumption of Village 0, which stabilizes in a few time steps.

\begin{figure}
    \centering
    \includegraphics[trim=0cm 0.1cm 0.1cm 0.1cm,clip,width=0.50\columnwidth,height=5.5cm]{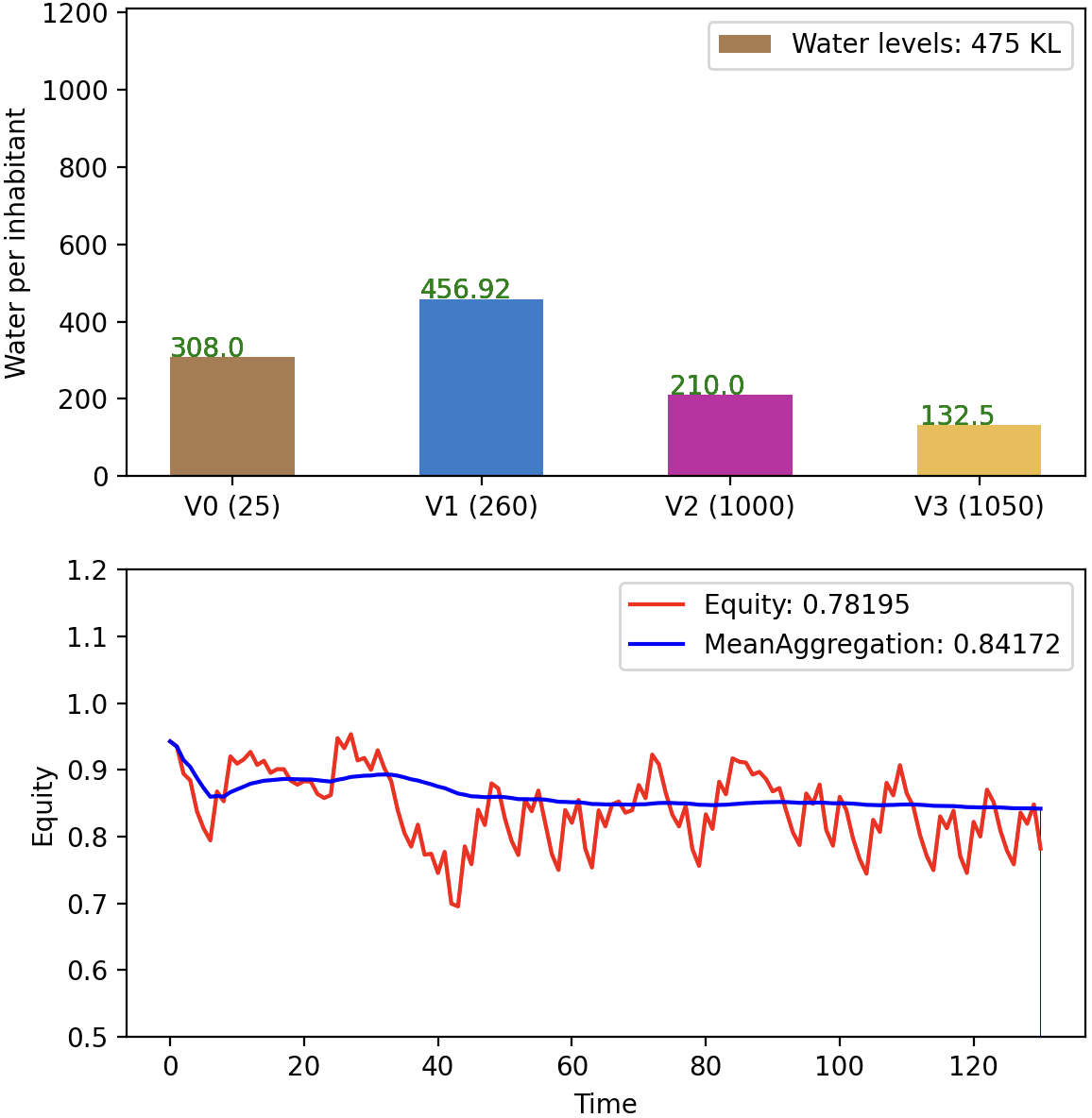}
    \includegraphics[trim=0.7cm 0.1cm 0.1cm 0.1cm,clip,width=0.49\columnwidth,height=5.5cm]{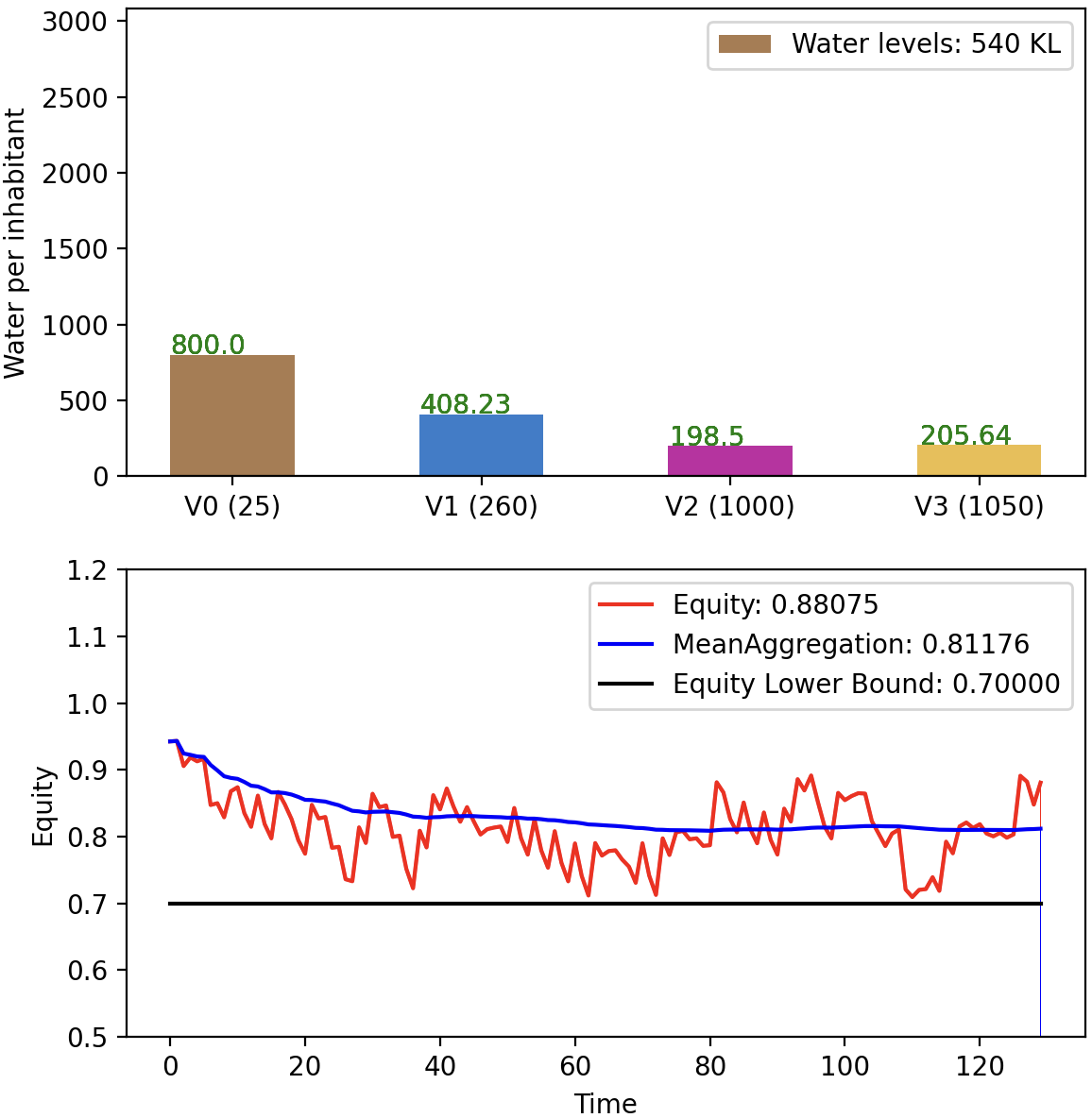}
    \vspace{-0.5cm}\caption{Performance of policies $\epsilon$-ADQL (left) and $\epsilon$-CADQL (right), with $\epsilon=0.1$ and $\tau=0.7$ marked at the horizontal black line. The upper diagrams show the distribution of water among the villages at the end of the episodes. The lower diagrams present the historic state equity (red) and aggregated average equity (blue).}\label{fig:epsilon01eadql}\label{fig:epsilon01ecadql}
\end{figure}

Figure~\ref{fig:epsilon01eadql} shows the results obtained with the two policies $\epsilon$-ADQL and $\epsilon$-CADQL,  learned with algorithms~\ref{alg:epsilon} and~\ref{alg:epsilon-constrained}. When comparing the ($\epsilon$-ADQL) strategy with the simple local policy, clearly, the $\epsilon$-local policy is able to improve the value-alignment of the whole path (e.g., the averaged historic equity/reward over all states of the path). Here, the obtained level is about $0.84$, versus $0.71$ of the local policy.

Comparing the obtained results to the $\tau$-constrained/$\epsilon$-local $\epsilon$-CADQL policy, we first see that the black line (the $\tau=0.7$ bound) is avoided.
In comparison, the local policy surpasses the bound many times and the $\epsilon$-ADQL policy does it once (approximately at time step 40). Additionally, the aggregated alignment -path average reward- of $\epsilon$-CADQL is similar to $\epsilon$-ADQL, albeit a bit lower: average reward of $0.812$ versus $0.842$. This behaviour seems logical, since in general, $\epsilon$-ADQL has more degrees of freedom for acting and thus, could find better policies regarding long-term value-alignment. Still, the $\epsilon$-CADQL policy is capable of bringing water to all villages, which is an advantage against the local policy. 

\begin{figure}
    \centering
    {\includegraphics[trim={0cm 0cm 0cm 0cm},clip,width=0.51\columnwidth, height=5.9cm]{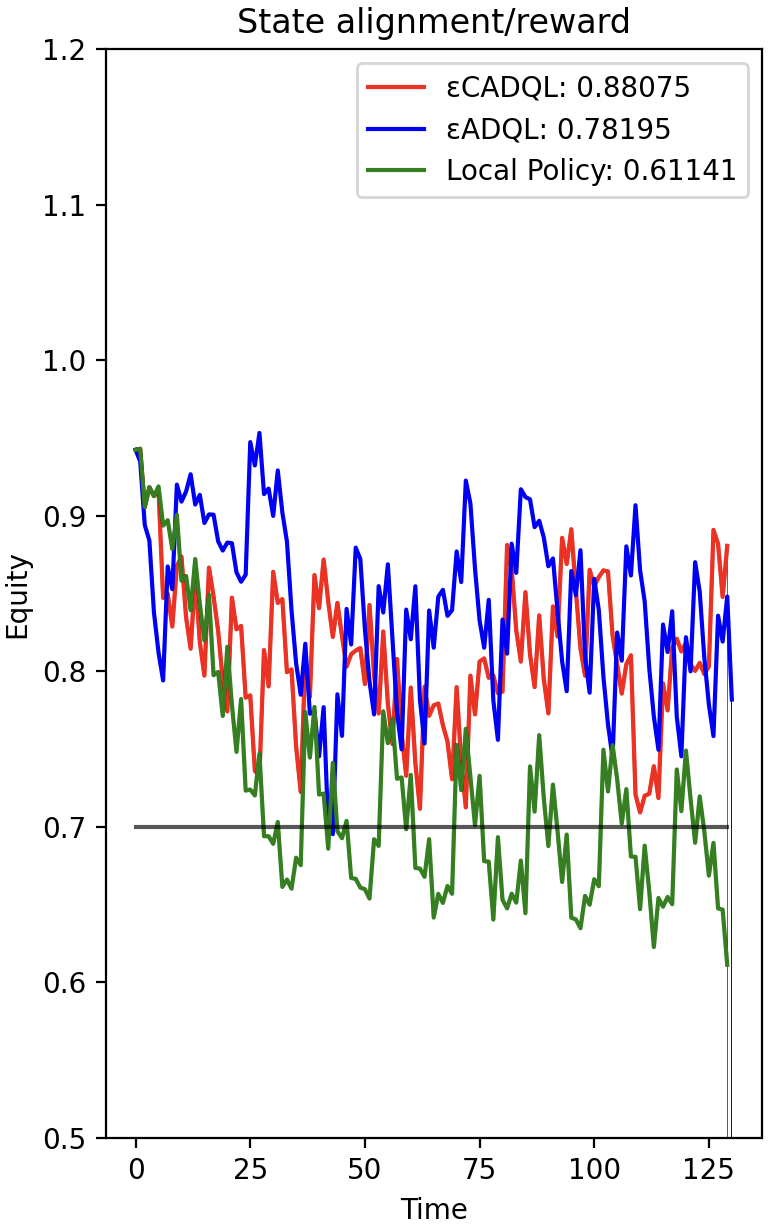}}
    {\includegraphics[trim={1.6cm 0cm 0.0cm 0cm},clip,width=0.44\columnwidth, height=5.9cm]{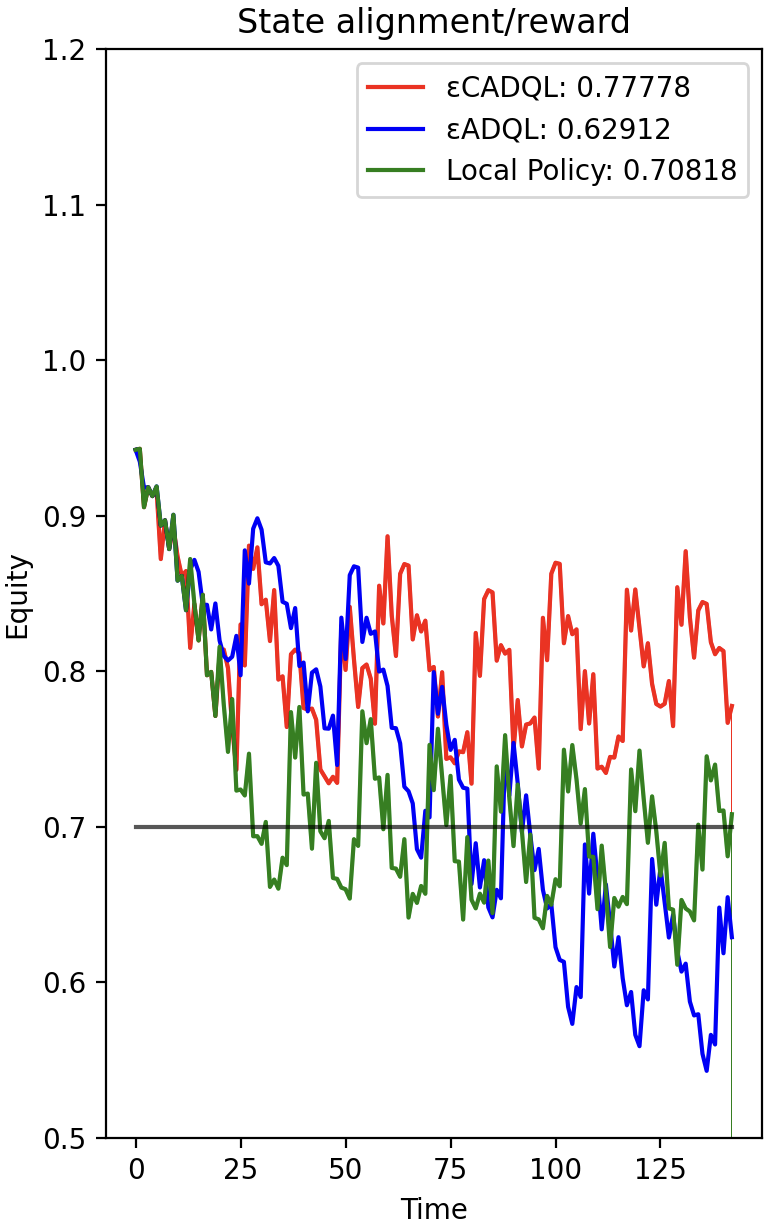}}
    \vspace{-0.5cm}
    \caption{Historic state equity ($f_{gini}$) for $\epsilon$-CADQL (red), $\epsilon$-ADQL (blue) and local policy (green). Left: results with $\epsilon=0.1$ and right: results with $\epsilon=0.01$ without retraining. }
    \label{fig:epsilon01}\label{fig:epsilon001}
\end{figure}

To have a clearer comparison between the three policies, we provide Figure~\ref{fig:epsilon01}. On the left side of this figure, we see the results when $\epsilon$ is set to $0.1$, which was already discussed for each individual policy. On the right side of Figure~\ref{fig:epsilon01}, we present the obtained results when applying a smaller $\epsilon$ ( $\epsilon=0.01$) in the evaluation. That is, the policies have been learned with $\epsilon=0.1$, but are then evaluated with the reduced $\epsilon$ of $0.01$, reducing the set of available actions greatly.

To see the relevance of the proposed training algorithms, a fourth ADQL (Average Double Q-Learning) policy was trained with $\epsilon$-ADQL (Algorithm~\ref{alg:epsilon}) but with $\epsilon=1$. We call it ``RL policy'' in our diagrams. Considering that our reward is bounded in $[0,1]$, this makes the training process equivalent to that of pure ADQL. The difference is in the sampling method, still done with Equation~\ref{eq:policy_sample}, to adhere to the $\epsilon$-local behaviour.

\begin{figure}
    \centering
    {\includegraphics[trim={0.1cm 0.2cm 0.25cm 0.4cm},clip,width=0.53\columnwidth, height=5.8cm]{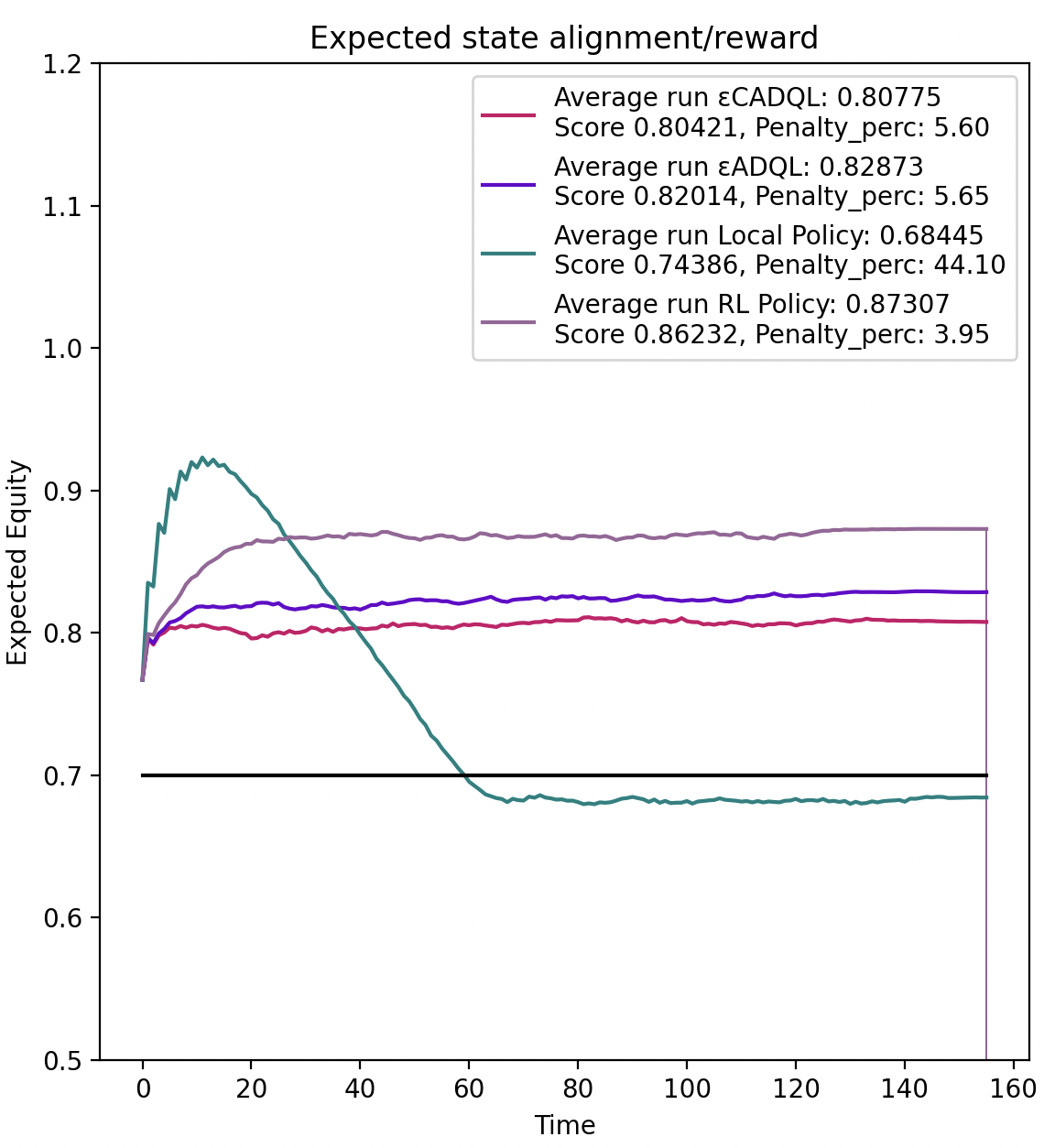}}
    {\includegraphics[trim={1.7cm 0.1cm 0.35cm 0.17cm},clip,width=0.46\columnwidth, height=5.77cm]{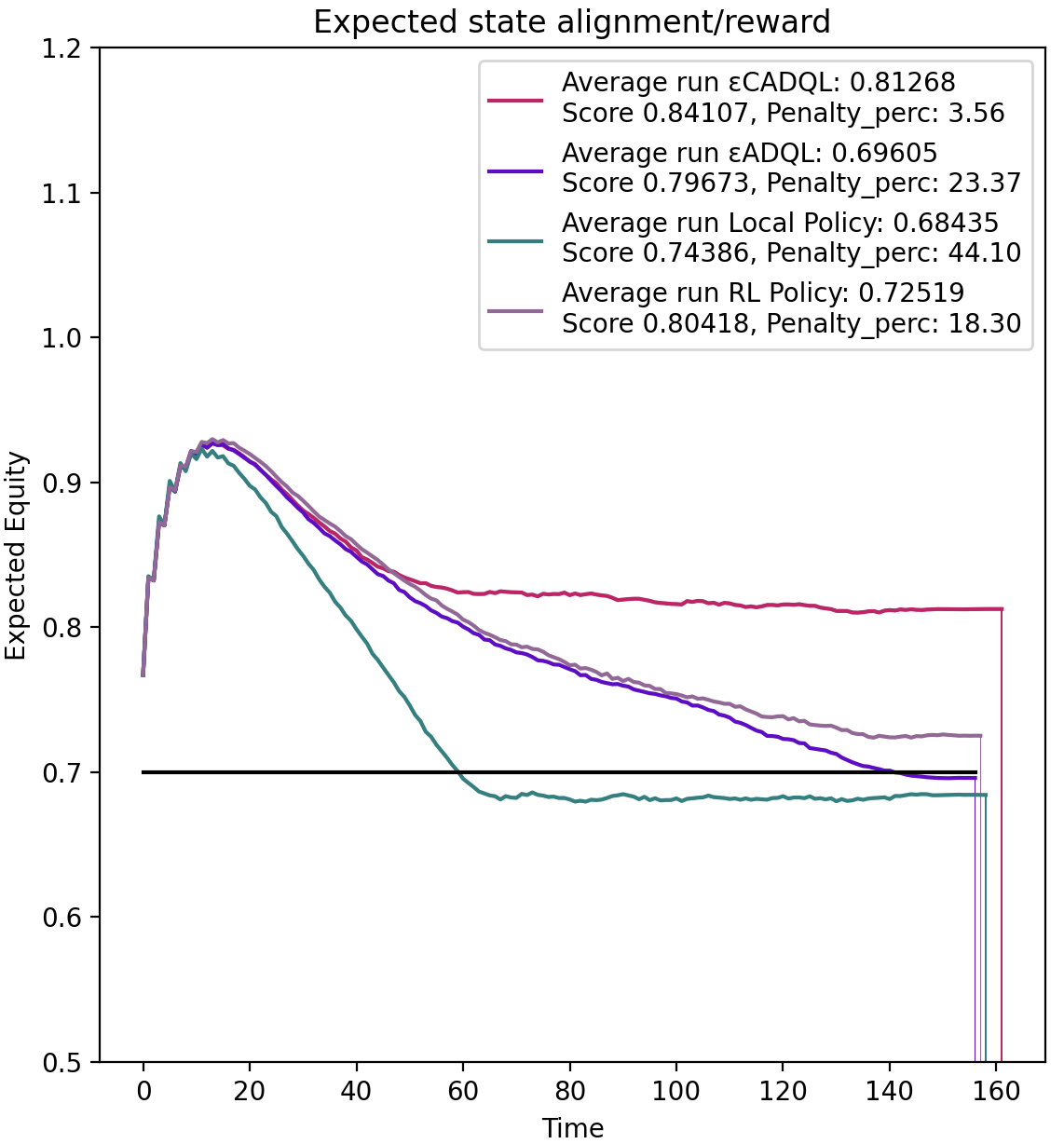}}
    \vspace{-0.5cm}
    \caption{Expected historic state equity ($f_{gini}$) for $\epsilon$-CADQL (dark red), $\epsilon$-ADQL (purple), local (dark green) and normal DQL (grey) policies. Left: results with $\epsilon=0.1$ and right: results with $\epsilon=0.01$. For each algorithm, the expected aggregated equity alignment is specified by the term \textit{Score} and the expected \textit{violation ratio} per time series length is given under \textit{Penalty\_perc}.}
    \label{fig:averages}
    
\end{figure}

Figure~\ref{fig:averages} shows the results of another experiment where we averaged the historic state equity/rewards of the paths obtained by the four policies from 1000 initial states sampled randomly with the process in Section~\ref{sec:training} with, again, $\epsilon = 0.1$ (left) and $\epsilon=0.01$ (right).\footnote{Different initial states and algorithms give different length paths. For visual purposes, the average historic reward time series seen in Figure~\ref{fig:averages} have been calculated by making all of them the same length, enlarging the shorter ones by repeating their ending rewards until getting as long as the longest series, which is previously cropped up to a maximum length of 1.2 times the default initial state experiment series. All the metrics, however, are calculated w.r.t. the original lengths and then averaged.}

The conclusions are the following. First, the three RL algorithms outperform in the mid/long-term the naively admissible local policy. Second, though $\epsilon$-ADQL achieves the best controlled performance\footnote{Unlike the ADQL policy, $\epsilon$-ADQL and $\epsilon$-CADQL are \textit{controlled} in the sense that both adhere to $\epsilon$-local behaviour during training.} under no changes in $\epsilon$, probably due to a the bigger state-space exploration, the ADQL policy is superior to the training-constrained proposals---it has a $3.65\%$ \textit{violation ratio} and an expected average reward (\textit{Score}) of $0.86$ versus $\approx5.6\%$ and $\approx0.82$ of our proposals, respectively---. Third, both the $\epsilon$-ADQL and ADQL policies fail in the mid-long term as their \textit{scores} drop below $0.81$ and their violation ratios grow to at least $18\%$ (suggesting a struggle for a sustained alignment); while $\epsilon$-CADQL will not, even diminishing its penalties to $3.56\%$ and increasing its score to $0.84$), showing $\tau$ constraint is avoided the intended way.

\section{Conclusions and Future Work}\label{sec:conclusions}

In this paper we have shown the importance of value-admissible behaviours for constructing value-aware decision-making systems. In previous work, these behaviours were proposed to enrich value-aligned goal-oriented decision-making, motivated by an equitable water distribution example. In this paper, we have revisited this scenario, introducing an infinite horizon distribution simulation. While adhering to two very general behaviours, namely the $\epsilon$-local behaviour (which relaxes an immediate need for short-term alignment) and the newly proposed $\tau$-constrained behaviour (which introduces "red line" constraints on possible states) we found more equitable paths in the long term than using the justifiable short-term focused local behaviour. We implemented two algorithms ($\epsilon$-ADQL and $\epsilon$-CADQL)
that use reinforcement learning to learn policies that implement the corresponding behaviours. In both cases we use an average reward setting, where rewards directly encode a value semantics function. We have proven the robustness and relevance of $\epsilon$-CADQL in particular for providing equity-safe solutions by combining aspects of the two admissible behaviours used.
We concluded that value-aware decision-making is possible with the learned policies combining complex value admissibility requisites.

As future work, note that the proposed algorithms need a special procedure for identifying the set of $\epsilon$-local admissible actions. Calculating it is time-consuming and will be infeasible in environments with a continuous action space. An alternative is to learn to identify the $\epsilon$-local admissible actions in constant time, e.g. with a discounted DQL algorithm with a small discount factor.  

Extending constraint value-aligned decision-making with \textit{value systems} is another open line of work, where complex value interactions should be represented not only by behaviours and state preferences but with explicable value taxonomies \cite{osman2023computational}.

\section*{Acknowledgements}
This work has been supported by grant VAE: TED2021-131295B-C33 funded by MCIN/AEI/ 10.13039/501100011033 and by the “European Union NextGenerationEU/PRTR”, by grant COSASS:
PID2021-123673OB-C32 funded by MCIN/AEI/
10.13039/501100011033 and by “ERDF A way of making
Europe”, and by the AGROBOTS Project of Universidad Rey
Juan Carlos funded by the Community of Madrid, Spain.

\balance

\bibliography{ia}

\end{document}